\documentclass[letterpaper, 10 pt, conference]{ieeeconf}  
\IEEEoverridecommandlockouts                              
\title{\LARGE \bf
Efficient Object Manipulation to an Arbitrary Goal Pose:\\Learning-based Anytime Prioritized Planning}

\author{Kechun Xu, Hongxiang Yu, Renlang Huang, Dashun Guo, Yue Wang, Rong Xiong
\thanks{{This work was supported by the National Nature Science Foundation of China under Grant 62173293}. Kechun Xu, Hongxiang Yu, Renlang Huang, Dashun Guo, Yue Wang, Rong Xiong are with the State Key Laboratory of Industrial Control Technology and Institute of Cyber-Systems and Control, Zhejiang University,
Hangzhou, China. Corresponding author,{\tt\small wangyue@iipc.zju.edu.cn}, Co-corresponding author, {\tt\small rxiong@zju.edu.cn}.}
}

\usepackage{color}
\usepackage{amsmath}
\usepackage{graphicx}
\usepackage{makecell}
\usepackage{booktabs}
\usepackage{gensymb}
\usepackage{hyperref}
\hypersetup{hidelinks}
\usepackage{booktabs}

\begin{document}

\maketitle
\thispagestyle{empty}
\pagestyle{empty}

\begin{abstract}

We focus on the task of object manipulation to an arbitrary goal pose, in which a robot is supposed to pick an assigned object to place at the goal position with a specific {orientation}. However, limited by the execution space of the manipulator with gripper, one-step picking, moving and releasing might be failed, where {a reorientation} object pose is required as a transition. In this paper, we propose a learning-driven \textit{anytime} \textit{prioritized} search-based solver to find a feasible solution with low path cost in a short time. In our work, the problem is formulated as a hierarchical learning problem, with the high level finding {a reorientation} object pose, and the low level planning paths between adjacent grasps. We learn an offline-training path cost estimator to predict approximate path planning costs, which serve as pseudo rewards to allow for pre-training the high-level planner without interacting with the simulator. To deal with the problem of distribution mismatch of the cost net and the actual execution cost space, a refined training stage is conducted with simulation interaction. A series of experiments carried out in simulation and real world indicate that our system can achieve better performances in the object manipulation task with less time and less cost.

\end{abstract}

\section{Introduction}

\begin{figure}[t]
\vspace{0.2cm}
  \centering
  \includegraphics[width=\linewidth]{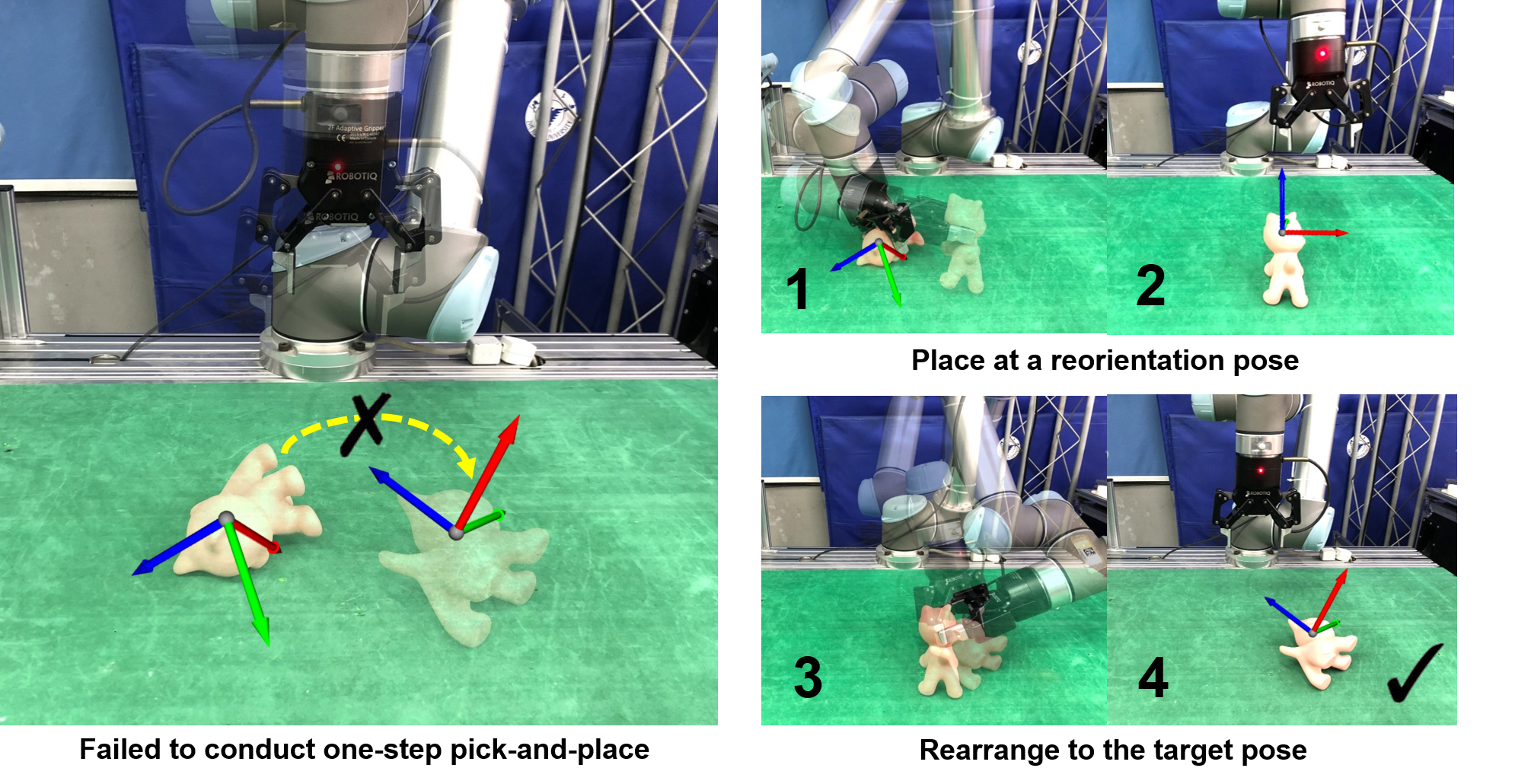}
  \vspace{-0.7cm}
  \caption{An example scenario (left) of the object manipulation task, where the robot is supposed to pick the cat model to a 6-DoF goal pose. In this case, one-step picking, moving and releasing will cause collision between the gripper and the table. By planning {a reorientation} object releasing pose, our system is capable of finding a collision-free path to reach the goal pose.}
  \label{problem}
  \vspace{-0.6cm}
\end{figure}

Object manipulation is essential for applications of embodied AI \cite{liu2021ocrtoc} such as table organization and supermarket checkout, where the robot is supposed to pick assigned objects to arbitrary 6-DoF goal poses. Finishing this task seems easy and efficient for human, but raises a challenge on the complexity of robotic planning. Note that in Fig. \ref{problem}, collision between the environment and the manipulator might sometimes hinder the simple one-step picking, moving and releasing. Instead, {a reorientation} object releasing pose is required so that the robot can re-plan the grasp poses in order to find a collision-free path to take the object to the goal pose. Therefore, the planning problem couples of the high-level {reorientation} object pose planning, as well as the low-level grasping and path planning, leading to a high-dimensional solution space and a very time-consuming search.

There are several works building systems for tabletop object rearrangement tasks \cite{garrett2020pddlstream}\cite{qureshi2021nerp}\cite{labbe2020monte}\cite{huang2019large}\cite{danielczuk2020object}. However, these works mostly focus on the task level of multi-object rearrangement, simplifying the object manipulation by assuming its optimality or success in default. Thus, these methods may fail when the assumption is invalid. Re-thinking about the human's highly efficient manipulation of an object to a goal pose, we consider that human generally does not follow an exact optimal plan of the task, but is inspired by some empirical approximations to a great extent. Imitating the way of human, to quickly find a sub-optimal but feasible plan for object manipulation, \cite{chai2020adaptive} proposes a RRT-like framework, which determines the {reorientation} object pose by minimizing the cost of the path planning. However, the method is limited in 2D plane with push actions. {Other works like \cite{ma2018regrasp}\cite{cheng2021learning}\cite{schmitt2019planning}\cite{wan2019preparatory}\cite{mitash2020task} also consider the possible failure of object manipulation by first placing the object to {a reorientation} pose utilizing the support of other objects, or making use of handover manipulation of dual arm}.

In this paper, we set to address the problem of object manipulation to an arbitrary 6-DoF goal pose by considering the full coupling of object pose, grasp pose and path. Following the human-like idea that finds a sub-optimal solution in a short time, we propose an efficient learning-driven \textit{anytime} \textit{prioritized} search-based solver. \textit{Anytime} means that our solver is supposed to provide a feasible solution quickly and optimize it over time, while \textit{prioritized} means that the solver follows some heuristics to find a better solution earlier. Thus, users can either access a good solution quickly, or wait for a better solution after some trials. {Note that our method focuses on single arm planning without support of other objects.}

As shown in Fig. \ref{fig:overview}, we formulate the problem as a hierarchical learning problem, which, in the high level, aims at finding {a reorientation} object pose, so that the low-level manipulator path between adjacent grasps is feasible and has minimum cost. Obviously, high-level planning calls for multiple times of running the low-level planning algorithm, whose complexity is thus exhaustive. Therefore, we train an estimator to predict the cost of the low-level plan without actually running the time-consuming low-level planning algorithm. Then, the estimated cost acts as a pseudo reward which allows for pre-training the high-level planner without interacting with the simulator. Finally, the whole hierarchical planner is refined in the simulator to decrease the error caused by the cost approximation during the pre-training. Thanks to the pre-training, this simulation based learning converges much faster. In the inference stage, the learned {reorientation} object pose planner prioritizes the search for a better solution in early trials. Experiments conducted in simulation and real-world environments demonstrate that our system can achieve better performances in the object manipulation task with less time and less cost. To summarize, the main contributions of this paper are:
\begin{itemize}
\item A hierarchical learning problem formulation for the task of object manipulation to an arbitrary goal pose.
\item A learning-driven \textit{anytime} \textit{prioritized} search-based solver for object manipulation to an arbitrary goal pose with high efficiency and better solution.
\item A pseudo reward guided pre-training based on the path cost estimator, which significantly accelerates the learning of {reorientation} pose planner.
\item The learned system is evaluated on both simulated and real world scenarios, of which the results validate the effectiveness.
\end{itemize}

\section{Related Work}

{
\subsection{Robotic Tabletop Rearrangement}
Recently, robotic tabletop rearrangement has become a topic of interest for researchers based on the studies of robotic grasping. In this task, the robot is supposed to pick objects and place them to their assigned poses \cite{liu2021ocrtoc}. \cite{zeng2018robotic} builds a pick-and-place system using a FCN network to predict grasps of both known and novel objects in cluttered environments. \cite{garrett2020pddlstream} formulates the rearrangement task in the symbolic domain and proposes an integrating symbolic planner to generate motion sequences. \cite{qureshi2021nerp} proposes a graph neural network approach for end-to-end rearrangement planning for unknown objects. However, such works \cite{garrett2020pddlstream}\cite{qureshi2021nerp}\cite{labbe2020monte}\cite{huang2019large}\cite{danielczuk2020object}\cite{zeng2018robotic}\cite{zeng2020transporter} mostly focus on high-level task planning, or grasp estimation, seldom considering the possible failure of the one-step object manipulation to the goal pose. 

\subsection{Object Reorientation and Regrasping}
There are also works explicitly considering the problem of object manipulation to a specific goal pose in tabletop rearrangement. \cite{newbury2021learning}\cite{jiang2012learning} learn object placement onto flat surfaces in upright orientations or in assigned constrained scenes. \cite{li2018push} encodes the physical information of an object to enable accurate push actions and generates {reorientation} object pose goals manually. \cite{chai2020adaptive} extends this work and determines the {reorientation} object poses by minimizing the cost of the path planning algorithm. \cite{ma2018regrasp}\cite{cheng2021learning} achieve desired regrasp poses by first placing objects to {reorientation} poses utilizing the support of other objects which form complex structure. \cite{schmitt2019planning}\cite{wan2019preparatory}\cite{mitash2020task} study regrasping planning for failed one-step object manipulation to constrained placements by utilizing handover manipulation of dual arm. 

Analogous to \cite{chai2020adaptive}, our method searches {a reorientation} object pose which minimizes the path planning cost. There are two main differences. First, \cite{chai2020adaptive} rearranges objects with push actions which limit the solution into 2D plane, while our method focuses on 6-DoF object manipulation. Second, the problem considered in this paper is coupled with three dimensions, which is more complex than the setting in \cite{chai2020adaptive}. Also, our work studies tabletop reorientation without support of complex structure with a single arm, which differs from the above works \cite{ma2018regrasp}\cite{cheng2021learning}\cite{schmitt2019planning}\cite{wan2019preparatory}\cite{mitash2020task}.
}

\section{Problem Statement}

We formulate the task of object manipulation to an goal pose as a multi-constraint optimization problem. Given the initial pose $p_0\in R^6$ and the goal pose $p_T\in R^6$, we define the grasp quality function as $h(p, g)\in(0,1]$, which demonstrates the grasp quality of the grasp pose $g\in R^6$ for the object pose $p\in R^6$, and the path planning cost function is defined as $m\left(g_{i}, g_{j}\right) \in(0, +\infty)$, where $g_{i}$ and $g_{j}$ represent the adjacent grasp poses. Then {a reorientation} object pose $p_I\in R^6$ is supposed to satisfy the following constraints:

\begin{equation}
\exists g_{0}, g_{I},\left\{\begin{array} { l }
{ h ( p _ { 0 } , g _ { 0 } ) > \alpha } \\
{ h ( p _ { I } , g _ { 0 } ^ { I } ) > \alpha } \\
{ h ( p _ { I } , g _ { I } ) > \alpha } \\
{ h ( p _ { T } , g _ { I } ^ { T } ) > \alpha }
\end{array} \text { and } \left\{\begin{array}{l}
m\left(g_{cur}, g_{0}\right)<+\infty \\
m\left(g_{0}, g_{0}^{I}\right)<+\infty \\
m\left(g_{0}^{I}, g_{I}\right)<+\infty \\
m\left(g_{I}, g_{I}^{T}\right)<+\infty
\end{array}\right.\right.\label{const}
\end{equation}
where $g_{cur}$ represents the current pose of the end effector, $g_{i}^{j}=p_{j} p_{i}^{-1} g_{i}$, and $\alpha$ is a threshold of grasp quality. If $p_I=p_T$ ({\it i.e.} $g_{0}^{I}=g_{0}^{T}, g_{I}=g_{T}$) satisfies above constraints, then the object can be directly manipulated without referring to {a reorientation} pose. Otherwise, {a reorientation} pose is needed as a transition for the completion of the object manipulation.

In this problem, the optimality of {a reorientation} pose lies in the feasibility and the cost of the whole planned manipulator path between adjacent grasps. Thus, the objective function of this optimization problem can be formulated as follows:
\begin{equation}
p_{I}^{*}, g_{0}^{*}, g_{I}^{*}=\operatorname{argmin}_{p_{I}, g_{0}, g_{I}} m\left(g_{0}, g_{I}^{T}\right)
\label{p1}
\end{equation}
where $m\left(g_{0}, g_{I}^{T}\right)=m\left(g_{cur}, g_{0}\right)+m\left(g_{0}, g_{0}^{I}\right)+m\left(g_{0}^{I}, g_{I}\right)+m\left(g_{I}, g_{I}^{T}\right)$. Note that the problem involves three dimensions: {reorientation} object pose planning of $p_I$, grasping planning of $g_0, g_I$, and path cost of $m(g_{0},g_I^T)$, which are coupled with one another, thus raising the difficulty for efficient search.

\section{Learning-based Hierarchical Planning}
\label{method}

\begin{figure*}[t]
  \centering
  \includegraphics[width=0.85\textwidth]{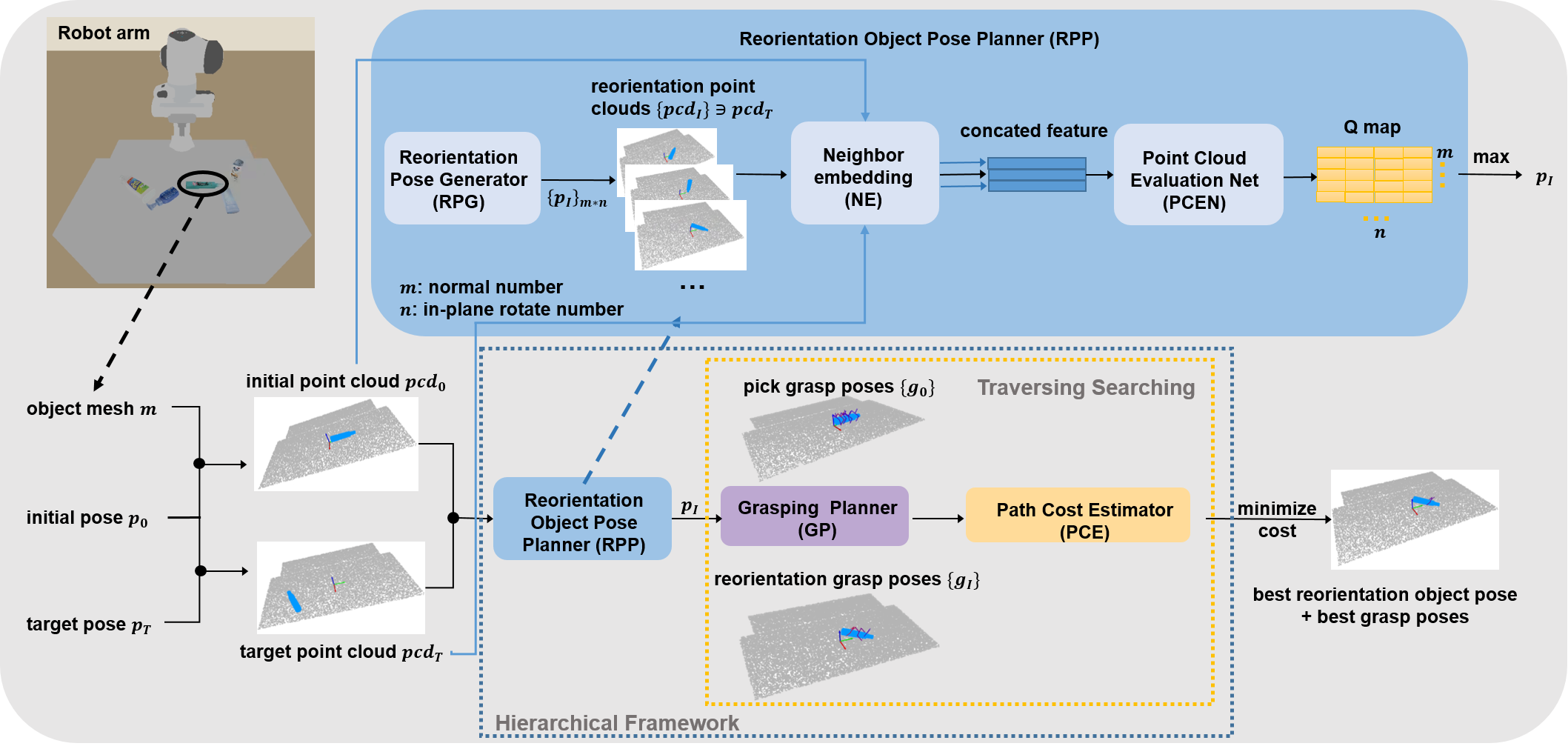}
  \vspace{-0.2cm}
  \caption{{\bf Overview} of our hierarchical framework. The {Reorientation} Pose Planner (RPP) is at the high level to distinguish the necessity of {a reorientation} object pose and generates one if needed. And the low level includes the grasping planner (GP) which conducts grasp detection and the Path Cost Estimator (PCE) which measures the planning cost in the whole manipulation process. Given an initial pose, a goal pose and an object mesh, our system generates corresponding point clouds $pcd_0$ and $pcd_T$ as the input of the RPP, which outputs a selected {reorientation} pose. If the selected {reorientation} pose is the goal pose, then the object will be directly manipulated to the goal pose using the grasp poses with the minimize planning cost. Otherwise, the object will be placed at the {reorientation} pose as a transition. }
  \label{fig:overview}
  \vspace{-0.6cm}
\end{figure*}

Obviously, traversing the whole space of the {reorientation} object poses and the grasp poses, and evaluating the corresponding path costs, can guarantee an optimal solution to (\ref{p1}). However, running a path planning algorithm exponential times takes lots of time, leading to the problem solving inefficient. Note that path planning is only related to the manipulator structure and the environment, and is independent of the {reorientation} pose and grasping. Thus, we can train a cost estimator to predict approximated path costs, denoted as $\hat{m}$, which avoids the actual running of a path planning algorithm, and largely improves the efficiency of high-level planning. Also, we adapt a pre-trained grasp model to generate prioritized grasp poses $G$ in a short time, which automatically satisfy validness constraints in (\ref{const}). In this way, the cost of a planned optimal {reorientation} object pose can be evaluated very efficiently as follows:
\begin{equation}
p_{I}^{*}, g_{0}^{*}, g_{I}^{*}=\operatorname{argmin}_{p_{I}, \{g_{0}, g_{I}\} \in G} \hat{m}\left(g_{0}, g_{I}^{T}\right)
\label{approx}
\end{equation}

We can surely employ exhaustive search to solve the problem, but it is still inefficient. Instead, we propose to learn a model to predict the {reorientation} object pose $p_{I}$ based on the object model as well as the initial and goal pose, which re-formulates (\ref{p1}) as a hierarchical learning problem.

To solve (\ref{approx}), as shown in Fig. \ref{fig:overview}, we build a hierarchical learning system that takes an initial pose, a goal pose and an object model as input, and outputs the $Q$-values of all {reorientation} object poses, as well as the corresponding grasps. Driven by prioritizing the object pose enumeration using the learned $Q$-values, an \textit{anytime} \textit{prioritized} search-based solver is implemented to provide a feasible solution quickly and optimize it gradually. To train the hierarchical learning system, we have three stages:

{\bf Stage \uppercase\expandafter{\romannumeral1}. Path Cost Estimator Training:} In this stage, we train an off-line cost net to estimate the path planning cost given the initial and goal pose, which can be regarded as an approximated environment feedback.

{\bf Stage \uppercase\expandafter{\romannumeral2}. {Reorientation} Object Pose Planner Training:} In this stage, we train {a reorientation} pose planner under the guidance of the cost net, which enables a feasible object manipulation path with less cost.

{\bf Stage \uppercase\expandafter{\romannumeral3}. Hierarchical Planner Refinement:} Considering the deviation between the pre-trained cost net approximation and the real execution cost, in this stage we further refine the {reorientation} pose planner and the path cost estimator by interacting with the simulator.

\subsection{Path Cost Estimator (PCE)}
\label{pce}

In the first stage, we train a network to get an approximate path planning cost $\hat{m}\left(g_{i}, g_{j}\right)$ given an initial pose $g_{i}$ and a goal pose $g_{j}$. We build a dataset by collecting samples of manipulator path planning and execution, each of which contains an initial pose $g_{i}$ and a goal pose $g_{j}$ for the end effector of the robot, and is labeled with the real cost $m\left(g_{i}, g_{j}\right)$ indicating the path length. Each initial pose and goal pose are concatenated as a 12-dim input, with 1-dim cost output. We randomly sample goal positions within the space above the table. And the robot plans and executes from the initial pose to the goal pose. The path planning algorithm is RRT. The labeled cost is numerically dependent on the number of path points which are sampled in a stationary interval.
\begin{equation}
m\left(g_{i}, g_{j}\right)=\left\{
\begin{array}{cl}
0.1\operatorname{path-len} \left(g_{i}, g_{j}\right), & \text { successful } \\
20, & \text { o.w. }
\end{array}\right.
\end{equation}

We train a MLP module to regress the path planning costs with a dataset containing 10k samples collected by a panda robotic arm with hand in SAPIEN \cite{Xiang_2020_SAPIEN}, a realistic and physics-rich robotics simulated environment.

\subsection{{Reorientation} Object Pose Planner (RPP)}
\label{RPP}
Our purpose of finding {a reorientation} object pose is to enable feasible path planning for the manipulator between adjacent grasp poses. And the optimal {reorientation} object pose and grasp poses are those that achieve the minimum cost of the whole object manipulation path. 

We apply DQN \cite{mnih2015human} to train our {reorientation} object pose planner {since it's hard to label {reorientation} poses for their feasibility}. For each episode, a model is randomly sampled from the model set, with an initial pose and a goal pose randomly sampled from the stable place poses (introduced in Sec.\ref{details}). Then the policy chooses an object pose $\hat{p}_I$ from a set of {reorientation} pose candidates including the goal pose. To define the reward, we first derive the minimum approximate cost sum of $\hat{p}_I$ by traversing the grasping as
\begin{equation}\label{mstarcost}
{g}^*_{0},{g}^*_{I} = \operatorname{argmin}_{{g}_{0}, {g}_{I} \in G} \hat{m}\left({g}_{0}, {g}_{I}^{T}\right)|_{p_I=\hat{p}_I}
\end{equation}
The minimum cost is denoted as $\hat{m}^*|_{p_I=\hat{p}_I}$, leading to the episode reward:
\begin{equation}
\text { episode reward }=5+\text { success }-\operatorname{cost} / 20
\end{equation}
\begin{equation}
\text { success }= \begin{cases}2, & \text { successful } \\ 0, & \text { o.w. }\end{cases}
\end{equation}

\begin{equation}
\label{cost}
\text { cost }= \begin{cases} \text {10$\hat{m}^*|_{p_I=\hat{p}_I}$}, & \text { successful } \\ 100, & \text { o.w. }\end{cases}
\end{equation}

{\bf Grasping planner (GP):} For the grasping planner, we adapt the graspnet in \cite{fang2020graspnet} with its provided pre-trained model, which predicts 6-DoF grasp poses with a scene point cloud as input. Also, graspnet yields few but accurate $G$, thus increasing the enumeration efficiency. In our system, we generate scene point clouds with known object models and a table model. For each object pose, a set of grasp pose candidates is obtained. These grasp poses are prioritized with grasp quality, which guarantees $h(p, g)>\alpha$, thus relaxing the constraints.

{\bf {Reorientation} object pose generator (RPG):} Our {reorientation} pose generation strategy is to select one among a set of pose candidates. We obtain some stable place poses $\left\{{p}_{I}\right\}$ against the plane table by calculating the probability that the model centroid falls in each triangle of the mesh. Since each triangle of the mesh corresponds to a face normal, poses in $\left\{{p}_{I}\right\}$ are with different normals. We sample $m$ normals from $\left\{{p}_{I}\right\}$ with augmentation of rotating about each normal by $n$ angles with a stationary step size, {thus generating $m\times n$ candidate {reorientation} poses}. Our {reorientation} pose candidate set is formed by these generated poses along with the goal pose.

For each pose in the {reorientation} pose candidate set, we generate a scene point cloud in which the known object is placed on the table with corresponding pose. Then we extract the point cloud feature through the network Neighbor Embedding (NE) \cite{Guo_2021} which encodes the input points into a new higher dimensional feature space. Each {reorientation} pose point cloud feature is concatenated with point cloud features of the initial and goal pose, and then is fed into Point Cloud Evaluation Net (PCEN) to get the predicted $Q$-value for this pose.

\subsection{Refinement}

In the second stage, our {reorientation} pose planner is pre-trained based on the PCE without any interaction with the simulator, which largely improves the sample efficiency. However, considering the data distribution deviation between the pre-trained PCE and the manipulator execution cost space, we add an extra stage to refine the PCE and RPP with data sampled from the interaction with the simulator {Note that PCE is supervised by the online real planning cost and RPP is trained in the same manner in Sec. \ref{RPP}}. By training the whole hierarchical planner in an end-to-end manner, we can further match the data distribution, as well as improve the performance, especially for RPP. Finally, the generated $Q$-value is employed to prioritize the enumeration based search for significant efficiency improvement.

\subsection{Implementation Details}
\label{details}

The MLP for the PCE contains two hidden layers with sizes [100, 10], which is trained with Adam optimizer and mse loss. As for RPP, our model set contains 90 {3D object} models, with sampled normal number $m=5$ and rotated angle number $n=6$. The stable place poses are obtained by computing the probability that the model centroid falls in each triangle of the mesh, which is realized in Trimesh \cite{trimesh}. {Apart from the goal pose, we constrain the $(x,y)$ tuples of the {reorientation} pose candidates to the same as the initial position, and compute the centroid heights of tabletop placement for $z$}. The network architecture of NE and PCEN are referred to PCT \cite{Guo_2021}. The architecture of NE is the same as that in \cite{Guo_2021}, while PCEN involves 4 stacked attention modules, followed by a convolution layer for fusion and two linear layers to generate the $Q$-value.

We train the networks with Adam optimizer (using fixed learning rates $10^{-4}$, weight decay $2^{-5}$, and betas $(0.9, 0.99)$) and Huber loss function. We use $\epsilon$-greedy as our exploration strategy, and $\epsilon$ is initialized as 0.5 then annealed to 0.1. Our future discount $\gamma$ is set as a constant at 0.9.

\section{Experimental Results}
\label{experiment}

\subsection{Simulation Experiments}
Our simulation environment involves a panda robotic arm with hand in SAPIEN.

\begin{table}[t]
\caption{COST PREDICTION RESULTS OF PCE}
\label{table:1}
\vspace{-0.2cm}
\centering
\begin{tabular}{cccc}
\toprule
& \makecell[c]{Success Rate/\%} & \makecell[c]{Average Error} & \makecell[c]{Average Time/ms}\\
\midrule
PCE & \makecell[c]{88.9} & \makecell[c]{5.67} & \makecell[c]{$1.18$}\\

Real Planning & $/$ & $/$ & \makecell[c]{$2.83\times10^3$}\\

\bottomrule
\end{tabular}
\vspace{-0.6cm}
\end{table}

The goals of the experiments are: 1) to test the performance of PCE. 2) to evaluate the effectiveness and efficiency of the solver. 3) to demonstrate that our system can successfully transfer from simulation to the real world. We compare the performance of our system to the following variants:

{\bf Exhaustive Solver (ES)} is an approach which finds the optimal {reorientation} pose from a candidate set by traversing searching, with the same {reorientation} pose candidate set determination approach with our method. For each {reorientation} pose and its corresponding grasp poses generated by graspnet, the robot plans and executes in the simulator and obtains the path planning cost by running the planning algorithm. For fair comparison, the graspnet and the simulator environment are the same as our method.

{\bf Exhaustive Solver with Cost Estimator (ESCE)} is an approach very analogous to {\bf ES}. The main difference is that this approach obtains approximate path planning costs from the cost estimator PCE without interaction with simulator.

\begin{figure}[t]
\vspace{0.1cm}
  \centering
  \includegraphics[width=0.8\linewidth]{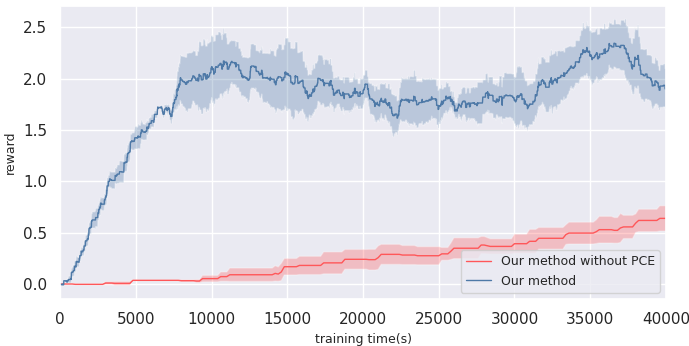}
  \vspace{-0.3cm}
  \caption{Comparing training performance of our method with an ablation method. {Note that {\bf Our method without PCE} will achieve a similar performance after much more training time, but here we only report 40000s to emphasize our efficiency.}}
  \label{fig:training}
  \vspace{-0.3cm}
\end{figure}

\begin{figure}[t]
  \centering
  \includegraphics[width=0.9\linewidth]{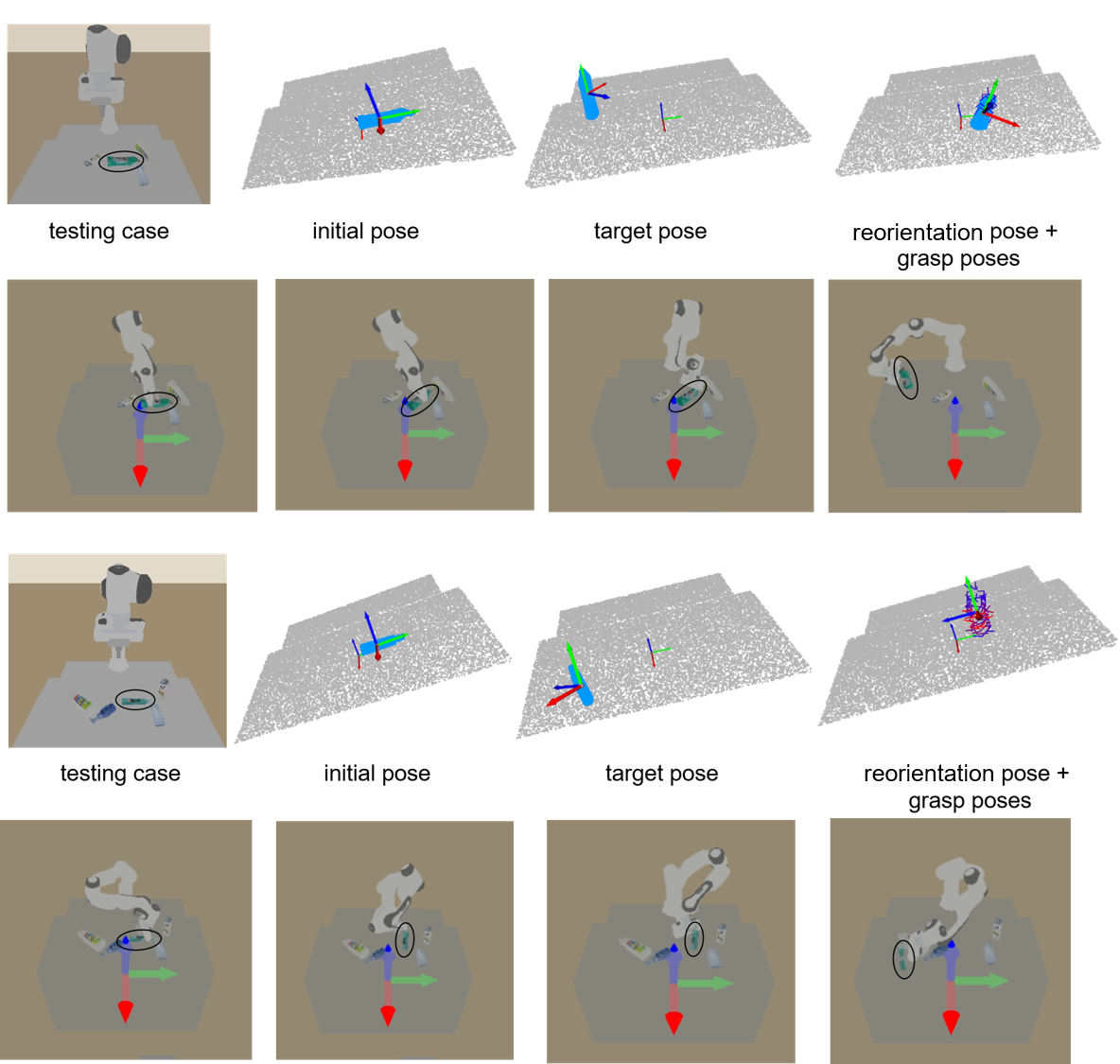}
  \vspace{-0.3cm}
  \caption{Example cases of planned {reorientation} object pose of RPP, where {a reorientation} object pose is needed due to the execution space limitation. The object pose is represented with a coordinate frame. The right of the figure shows the selected {reorientation} object pose with its grasp pose candidates.}
  \vspace{-0.7cm}
  \label{fig:case_ip}
\end{figure}

\begin{figure}[t]
  \centering
  \includegraphics[width=0.9\linewidth]{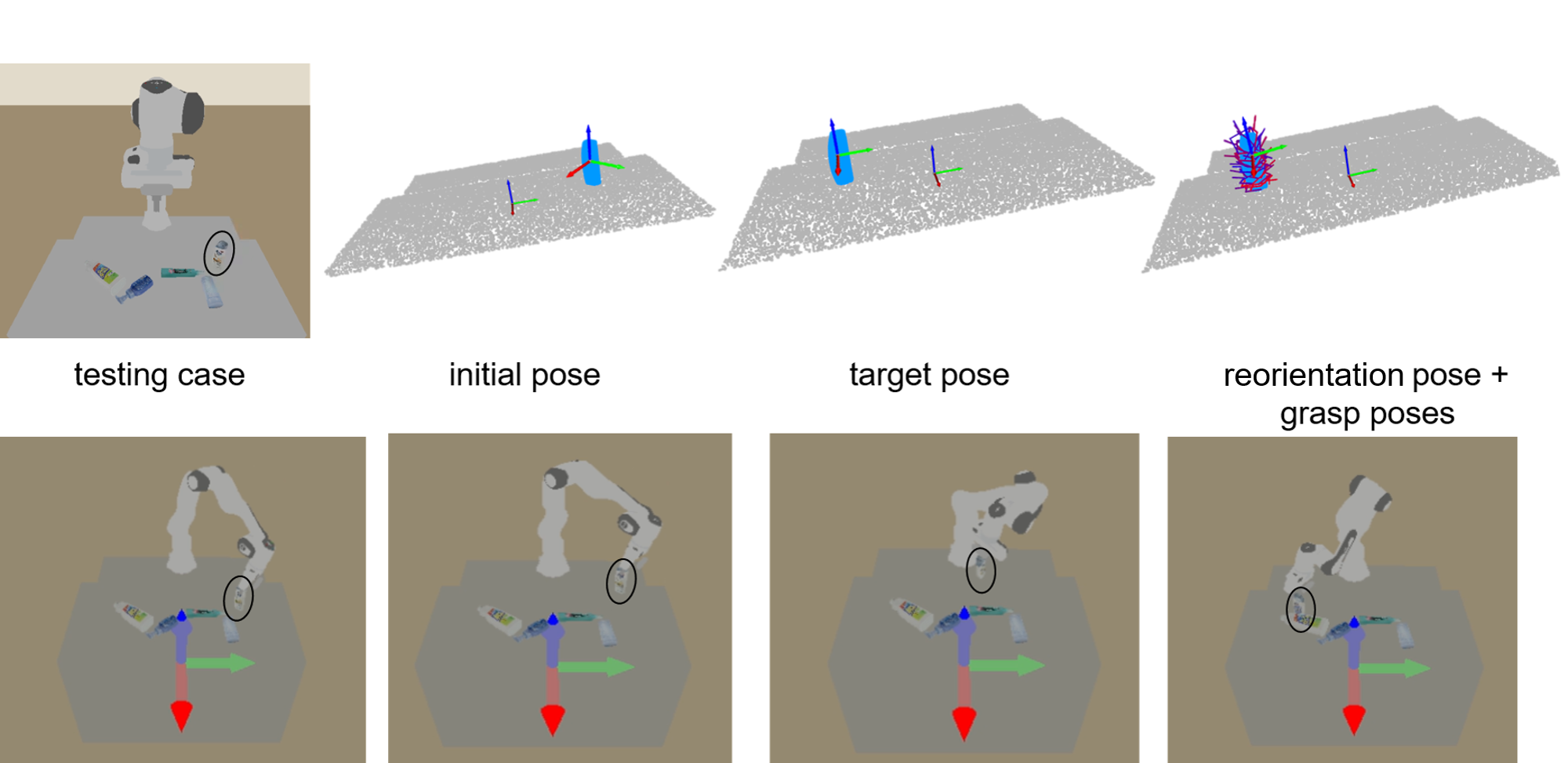}
  \vspace{-0.3cm}
  \caption{An example case of planned {reorientation} object pose of RPP, where the direct manipulation to the goal pose can be achieved. The object pose is represented with a coordinate frame. In this case, RPP chooses the goal pose as the next pose, and omits the {reorientation} transition process.}
  \label{fig:case_wo_ip}
  \vspace{-0.2cm}
\end{figure}

\begin{figure}[t]
  \centering
  \includegraphics[width=0.8\linewidth]{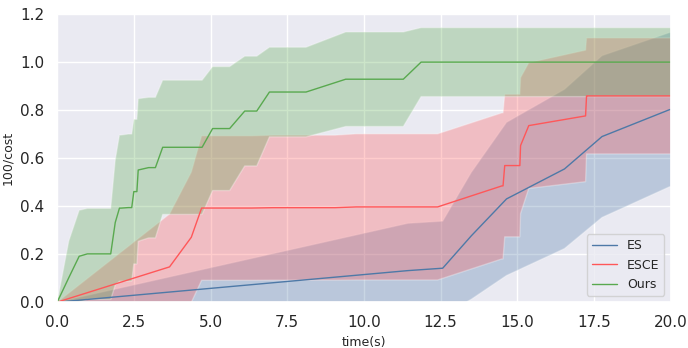}
  \vspace{-0.3cm}
  \caption{Planning cost sum curves over planning time of all methods. The performance difference can be intuitively seen from the margin area between the curves. {\bf ES} costs the longest planning time to plan, while our method can obtain a sub-optimal solution in a short time interval and continuously optimize it over time. {\bf ESCE} can also achieve a high planning efficiency, with a slower improvement of performance in the early time compared to our method.}
  \label{fig:baseline}
  \vspace{-0.6cm}
\end{figure}

\begin{figure*}[t]
\vspace{0.2cm}
  \centering
  \includegraphics[width=0.8\textwidth]{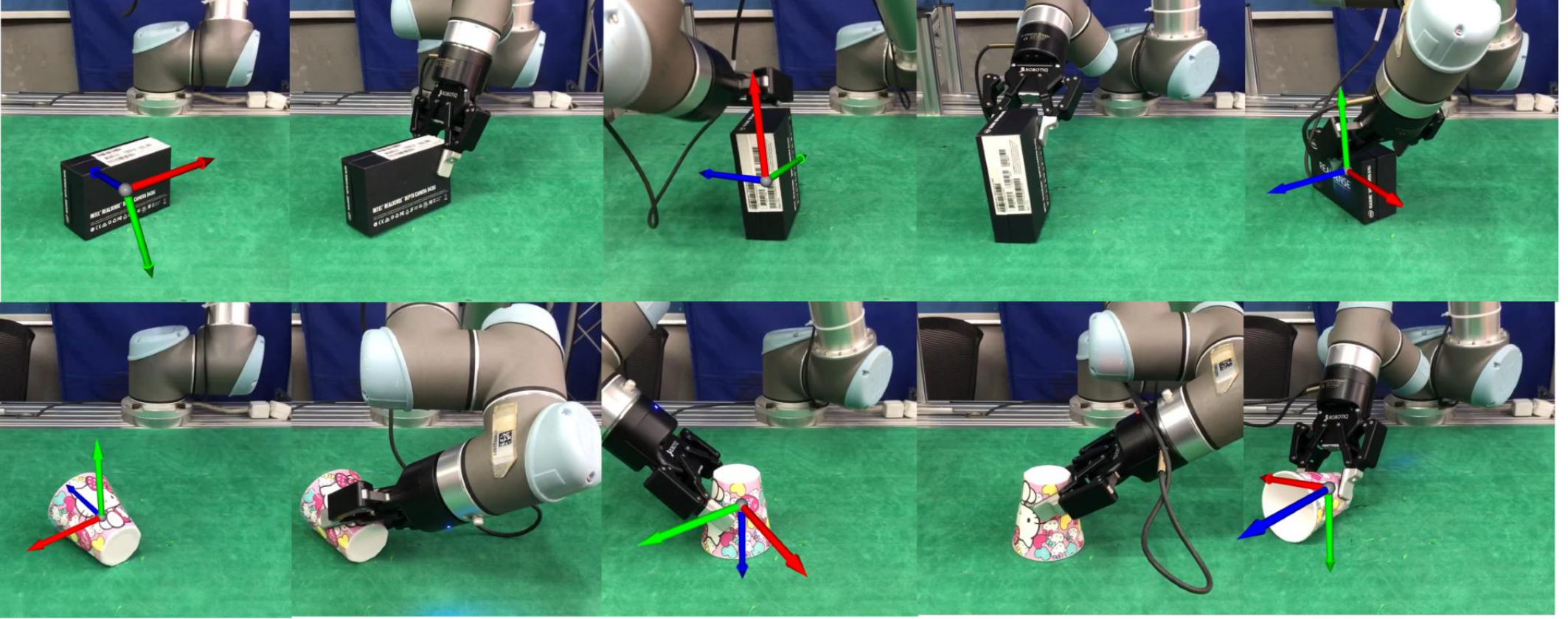}
  \vspace{-0.1cm}
  \caption{Example action sequences in real-world experiments. One-step rearrangement is hindered by the execution space of the manipulator. To complete this rearrangement process, our method generates {a reorientation} object pose. The object is firstly placed at this pose as a transition and then transferred to the goal pose.}
  \label{fig:action-sequence}
  \vspace{-0.7cm}
\end{figure*}

{\bf Cost Prediction Results:} In this experiment, we test the performance of PCE. We collect 2.5k path planning testing samples in simulation environment by randomly assigning initial and goal poses for the end effector of the manipulator. Details of data collection are the same as Sec. \ref{pce}. If our cost estimator is capable of correctly distinguishing a sample between successful and failed planning, then that test is considered successful. Results of success prediction rate, average cost prediction error and average prediction time are reported in Table \ref{table:1}, which suggest that our cost estimator is capable of well distinguishing whether there exists a feasible trajectory for the object manipulation from the initial to the goal with much higher efficiency than real planning. As for the planning cost, since the path planning algorithm RRT generates paths with randomness, the planning cost is different each time. Thus, the average error metric is in an acceptable range. {Note that real planning is regarded as ground truth for our estimator to calculate success rate and average error. Therefore, there are no report of these two metrics for real planning.}

{\bf Training Efficiency:} We compare our method with an ablation method to test whether training under the guidance of PCE can improve sample efficiency. The ablation method is similar to our method, except for training without PCE (\textit{i.e.} with real samples by interacting with the simulator).

We record rewards versus training time to indicate performance during the training process and present the curves in Fig. \ref{fig:training}. It's obvious that our training approach can improve the performance with a faster pace, which indicates a higher training efficiency. By utilizing the pre-trained cost net, our training pipeline can omit the real path planning and execution, thus obtaining pseudo environment rewards quickly.

{\bf Case Studies:} We report cases to qualitatively demonstrate the performance of RPP. As shown in Fig. \ref{fig:case_ip}, the direct manipulation to goal pose of ``repellent'' {(one of the object models used in our experiments)} will be failed due to the limitation of the execution space of the manipulator. In this case, RPP will plan an appropriate {reorientation} pose. By standing the ``repellent'' first, the robot can easily rotate it to the goal pose, thus completing the whole rearrangement process. For cases like Fig. \ref{fig:case_wo_ip}, where the object can be directly manipulated to the goal pose, our RPP will predict the goal pose as the next object pose, which omits the {reorientation} transition process.

{\bf Single-time Planning:} As a whole system, we compare our solver with baselines with 10 testing cases containing 10 different objects (with known models) in unseen scenarios. For each testing case, the robot should rearrange the assigned object from the assigned initial pose to the goal pose in a scenario containing at least 5 objects.

In this experiment, we conduct a single time planning of each testing case for every method, and record the average planning time, success rate and real planning cost sum in Table \ref{table:2}. If the object manipulation process is completed without planning failure, then the test is considered successful. {And the cost metric is the average real path planning cost sum $10{m}\left(g_{cur}, g_{I}^{T}\right)$ of the whole pick-and-place process for successful manipulations}. As shown in the table, {\bf ES} achieves the highest success rate and lowest cost sum. This is because {\bf ES} conducts traversing searching with real path planning cost, which ensures the optimal solution of minimum path planning cost. Thus, {\bf ES} can serve as a reference for other two methods. Based on the cost estimator, other two methods can achieve much higher efficiency for planning, but with lower success rate and higher average cost sum. Compared to {\bf ESCE}, our method can achieve higher success rate with much less time and less cost, which confirms that our RPP can effectively improve the planning efficiency and our refinement training can further match the cost net distribution and improve the overall performance. In a word, our method can obtain a better solution in a short time.

{\bf Online Planning:} Following the same experimental setting in single-time planning testing, we then test the performance for online planning. All solvers plan online for each testing case. Once a solution is generated, the robot will execute and get a planning cost sum. If the solver hasn't generated a solution till the current time or the generated solution is failed, then the cost is set as $+\infty$. Also, as time goes on, the solver will update its solution with the best one with the lowest cost. In this experiment, the solver performance is measured by average planning time and average path planning cost, which are key metrics for anytime prioritized planner. To be more intuitive, we record the reciprocal of the planning cost sum over planning time for all objects and demonstrate the average curves with standard deviation in Fig. \ref{fig:baseline}. It can be seen in the figure that our method outperforms all baselines in planning time. Although {\bf ES} can ensure an optimal solution, it costs the longest time to plan. In constrast, our method can obtain a sub-optimal solution in a short time interval and continuously optimize it over time, finally reaching an optimal solution. {\bf ESCE} can also achieve a higher planning efficiency, but in the early time the improvement of performance is slower than our method.

\begin{table}[h]
\vspace{-0.2cm}
\caption{SIMULATION RESULTS FOR SINGLE PLANNING}
\label{table:2}
\vspace{-0.2cm}
\centering
\begin{tabular}{cccc}
\toprule
\makecell[c]{Method} & \makecell[c]{Success/\%} & \makecell[c]{Time/s} & \makecell[c]{Cost} \\
\midrule
\makecell[c]{ES} & \makecell[c]{{100.0}} & \makecell[c]{155} & \makecell[c]{{96}} \vspace{0.05cm}\\
\hline \vspace{-0.2cm}\\
\makecell[c]{ESCE} & \makecell[c]{80.0} & \makecell[c]{17.6} &
\makecell[c]{153}\\

\makecell[c]{Ours} & \makecell[c]{{\bf 90.0}} & \makecell[c]{{\bf 2.27
}} & \makecell[c]{\bf 132}\\
\bottomrule
\end{tabular}
\vspace{-0.5cm}
\end{table}

\subsection{Qualitative Real-World Experiments}

In this section, we qualitatively test our system in real-world experiments with several cases. To obtain the initial pose and the goal pose, each object is placed at the corresponding poses separately, and the manipulator takes multi-view images and projects the model with the camera extrinsics from each view. Then poses are manually adjusted based on the initial values from feature matching. 

Two example action sequences are shown in Fig. \ref{fig:action-sequence}, which suggest that our system can be successfully transferred to the real world. For the realsense box, one-step turning over will lead to collision. Our system complete this manipulation task by standing the box first to enable a feasible trajectory to the goal pose. As for the hello kitty cup, {a reorientation} pose is also applied for a low cost manipulation path. Overall, our system can successfully apply to some cases in the real world, which confirms the effectiveness of our method. 

\section{Conclusion}

In this work, we study the task of object manipulation to goal pose, which is formulated as a multi-constraint optimization problem. We propose an efficient learning-driven {\it anytime prioritized} search-based solver, and build a hierarchical learning framework with three stages of training. Simulation experiments show that our method can achieve better performance with less planning time. Real-world experiments suggest our system is capable of effectively applying to the real world. It is worth noting that our method can be easily extended to multi-step (\textgreater 2) object manipulation problem and can serve as a front work of rearrangement tasks.



\newpage
\bibliographystyle{IEEEtran}
\bibliography{IEEEabrv,ref}

\end{document}